\documentclass{article}
\usepackage{ijcai22}

\usepackage{adjustbox}
\usepackage{multirow}

\usepackage{graphicx}
\usepackage{subfigure}

\usepackage{times}
\usepackage{soul}
\usepackage{url}
\usepackage[hidelinks]{hyperref}
\usepackage[utf8]{inputenc}
\usepackage[small]{caption}
\usepackage{graphicx}
\usepackage{amsmath}
\usepackage{amsthm}
\usepackage{booktabs}
\usepackage{algorithm}
\usepackage{algorithmic}
\usepackage{amsfonts}
\title{RAPQ: Rescuing Accuracy for Power-of-Two Low-bit Post-training Quantization}

\author{
Hongyi Yao
\and
Pu Li\and
Jian Cao\footnote{Corresponding author}\and
Xiangcheng Liu\and
Chenying Xie\And
Bingzhang Wang
\affiliations
Peking University
\emails
yhy@stu.pku.edu.cn,
spurslipu@pku.edu.cn,
caojian@ss.pku.edu.cn,
liuxiangcheng@stu.pku.edu.cn,
402600293@qq.com,
13919334117@163.com
}

\begin{document}

\maketitle

\begin{abstract}
We introduce a Power-of-Two low-bit post-training quantization(PTQ) method for deep neural network that meets hardware requirements and does not call for long-time retraining. Power-of-Two quantization can convert the multiplication introduced by quantization and dequantization to bit-shift that is adopted by many efficient accelerators. However, the Power-of-Two scale factors have fewer candidate values, which leads to more rounding or clipping errors. We propose a novel Power-of-Two PTQ framework, dubbed RAPQ, which dynamically adjusts the Power-of-Two scales of the whole network instead of statically determining them layer by layer. It can theoretically trade off the rounding error and clipping error of the whole network. Meanwhile, the reconstruction method in RAPQ is based on the BN information of every unit. Extensive experiments on ImageNet prove the excellent performance of our proposed method. Without bells and whistles, RAPQ can reach accuracy of 65\% and 48\% on ResNet-18 and MobileNetV2 respectively with  weight INT2 activation INT4. We are the first to propose the more constrained but hardware-friendly Power-of-Two quantization scheme for low-bit PTQ specially and prove that it can achieve nearly the same accuracy as SOTA PTQ method. The code\footnote{https://github.com/BillAmihom/RAPQ} was released.
\end{abstract}

\section{Introduction}

In recent years, convolutional neural network (CNN) has been widely used in computer vision tasks. The improvement of hardware computation considerably accelerates model evolution, which produces deeper and more complex CNN models to pursue even higher accuracy. However, the deep CNN models are difficult to be deployed on resource-limited edge devices. How to reduce the model scale while maintaining the model accuracy is a trending topic in current research. In this paper, we study quantization which aims to reduce bit-width of weights and activations to enable fixed-point computation and less memory space.

Based on data usage, model quantization can be divided into three categories: (1) quantization-aware-training (QAT), (2) post-training quantization (PTQ) and (3) data-free quantization (DFQ). QAT requires fine-tuning the model on the whole dataset, which inevitably requires a large amount of GPU resources and time cost. In contrast, PTQ demands only a small set of readily available calibration data. Although DFQ achieves quantization without dataset, its accuracy has not reached the desired level and it is difficult to apply in industrial scenarios. Therefore, this paper focuses on improving PTQ performance.

Moreover, scale factors that are constrained to the form of Power-of-Two make quantization and dequantization convert to simple bit-shift. However, compared with the float scale factors,the Power-of-Two value is essentially a discrete approximation of the float value, which will cause more rounding error or more clipping error. This significantly reduces performance of the quantized model.

We are the first to implement hardware-friendly Power-of-Two low-bit PTQ and surprisingly observe that constrained Power-of-Two PTQ can achieve nearly the same accuracy as SOTA PTQ method.

\section{Related Work}
\subsection{Network Quantization}
 QAT~\cite{krishnamoorthi2018quantizing} mainly adopts the STE for gradients approximation to solve the non-differentiable round problem.  ~\cite{gong2019differentiable} uses a differentiable quantizer to gradually approach the round function. However, QAT usually depends on the whole dataset and GPU resources to train the model. Without high cost, most models can be safely quantized to 8-bit even lower-bit by PTQ. AdaRound ~\cite{nagel2020up} proposed to learn the rounding way by layer reconstruction brings much improvement of accuracy. BRECQ ~\cite{li2020brecq} focused more on block reconstruction with better accuracy at 2-bit weight quantization than AdaRound.

The Power-of-Two scale factor~\cite{miyashita2016convolutional} has the advantage of reducing computation complexity. But it meanwhile produces accuracy loss. To solve this problem, ~\cite{li2019additive} proposed an efficient method APoT for the weights and activations with bell-shaped and long-tailed distribution in neural networks. But APoT is a non-uniform quantization QAT scheme.

\section{Motivation}

For simplicity, we omit the analysis of $\boldsymbol {bias}$ as it can be merged into activation. In this way, the forward propagation of CNN can be expressed by Equation (\ref{equ1}).
% \begin{align}

\begin{equation}
\boldsymbol{x}_{(k+1)}=\mathcal{R}\left(\boldsymbol{y}_{(k)}\right) = \mathcal{R}\left(\boldsymbol{W}_{(k)} * \boldsymbol{x}_{(k)}\right)
\label{equ1}
\end{equation}
% \end{align}
\\
where $*$ represents the convolution. $\mathcal{R}\left(\boldsymbol\cdot\right)$ is the activation function. $\boldsymbol{x}_{(k)}$ is the input of the ${k}$-th layer, $\boldsymbol{x}_{(k+1)}$ is the output activation of the kth layer, and $\boldsymbol{y}_{(k)}$ is the convolution result of the ${k}$-th layer.

The essence of quantization is to map floating-point number to low-bit fixed-point number. The quantization and dequantization process of non-uniform quantization often brings huge computing burden. So in this paper, we focus on uniform affine quantization. To quantize vector $\boldsymbol{x}$, with ${s}$ denoting the scale of the floating-point number mapped to the fixed-point number, and ${z}$ denoting the zero-point of shifting the number to the specified range, the process of quantization of input $\boldsymbol{x}$ can be described by Equation (\ref{equ2}).
\begin{equation}
\boldsymbol{\hat{x}}=s \cdot
\left[clip
\left(\left\lfloor\frac{\boldsymbol{x}}{s}+z \right\rceil, n, p\right)-z\right]
\label{equ2}
\end{equation}\\where $\left\lfloor\cdot\right\rceil$ represents the rounding. The quantized variables are marked by $\hat{}$ , ${clip}(\cdot)$ represents that input will be clipped into $[n, p]$, in the case of asymmetric quantization $n=0$ and $p=2^{b}-1$, b is the bit-width.

\begin{figure}[t]
\centering
\includegraphics[width=0.9\columnwidth]{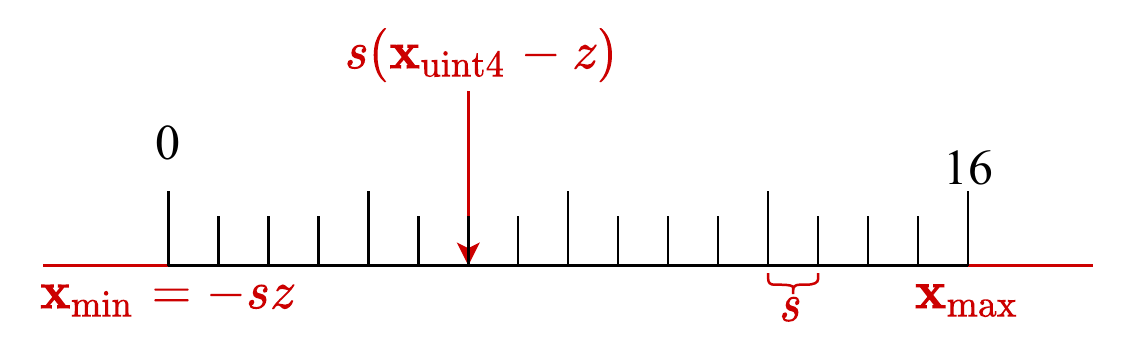} % Reduce the figure size so that it is slightly narrower than the column. Don't use precise values for figure width.This setup will avoid overfull boxes.
%0502\vspace{-2mm}
\caption{Visual explanation of the asymmetric uniform affine quantization grids for a bit-width of 4.}
\label{fig1}
\end{figure}

If we use a grid diagram to represent the range of fixed-point numbers, we can interpret ${s}$ as the step length between two grids, and ${z}$ determines the utilization interval of fixed-point numbers\cite{nagel2021white}, as shown in Figure \ref{fig1}.

\begin{figure}[htbp]
\centering
%0502\vspace{-1mm}
\subfigure[Data quantied by ${\lfloor\log _{2} s\rfloor}$]
{
\includegraphics[width=4cm]{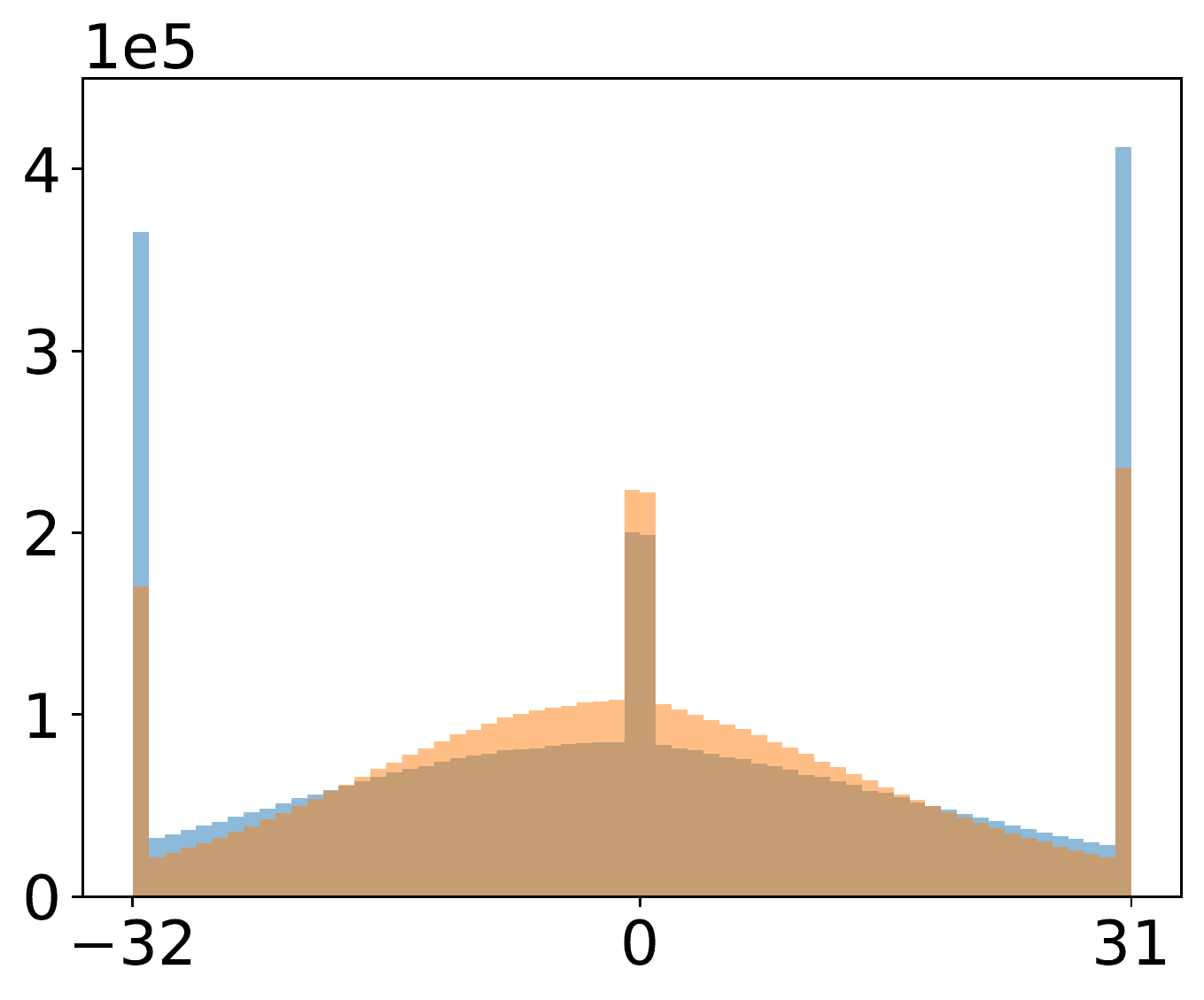}
\label{fig2(a)}
%\caption{fig1}
   }
%0502\vspace{-1mm}
\subfigure[Data quantied by ${\lceil\log _{2} s\rceil}$]
{
\includegraphics[width=4cm]{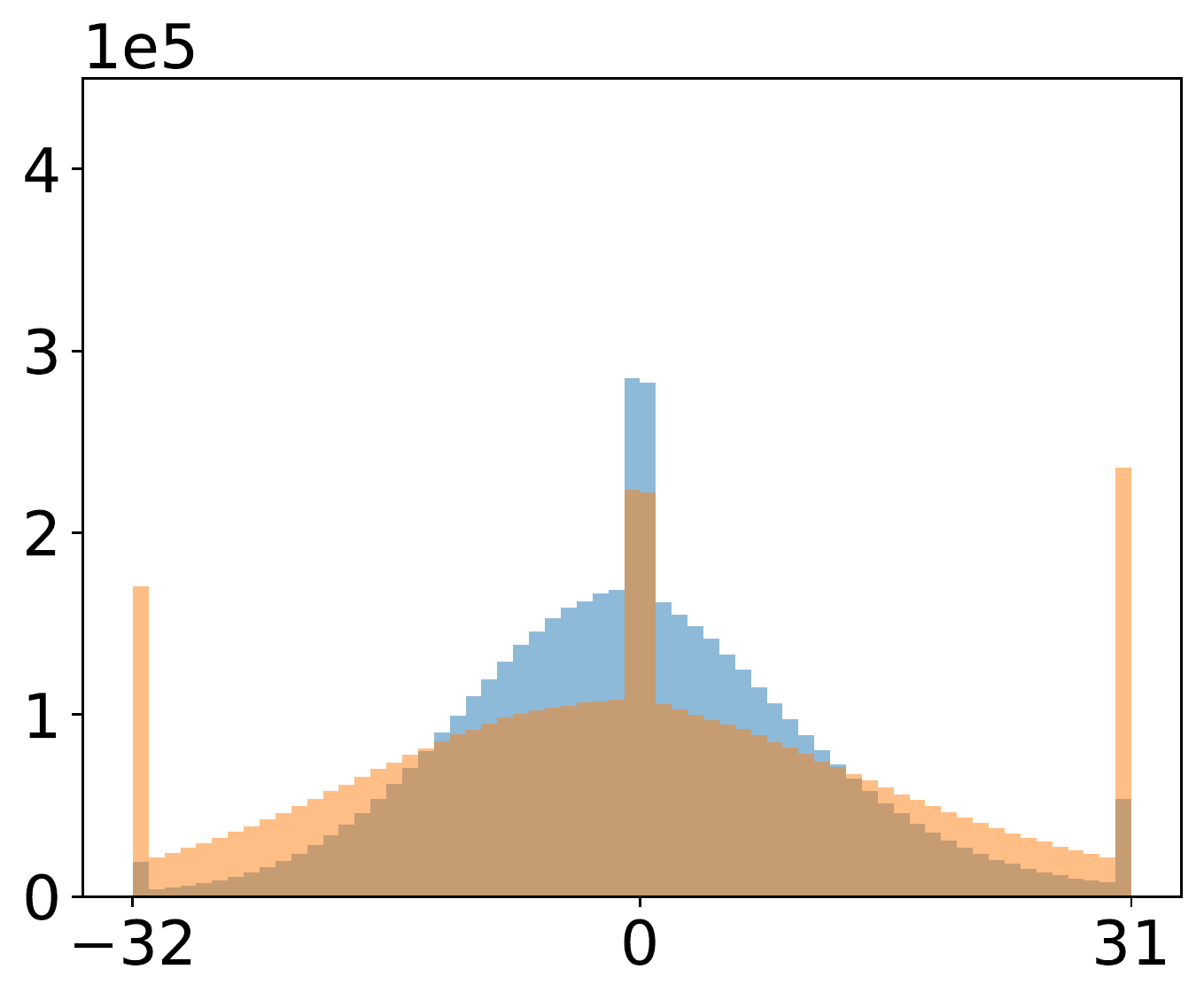}
\label{fig2(b)}
}
%0502\vspace{-3mm}
\caption{6-bit symmetric uniform affine quantization for weights in 73rd layer of DarkNet53}
\label{fig2}
\end{figure}

\subsection{Relationship Between Scale and Rounding Error {\&} Clipping Error}
So far, there is no Power-of-Two quantization framework specifically for low-bit PTQ. But the hardware requirement for Power-of-Two scale factors usually occurs.
The naive method is to first quantize the model with float scale factors, and replace them with the closest Power-of-Two scale factors. Expressed in Equation (\ref{equ3}) is 
\begin{equation}
\boldsymbol{\hat{x}}=s_{pow-2}\cdot\left[\operatorname{clip}\left(\left \lfloor \frac{\boldsymbol{x}}{s_{pow-2}}+z^{\prime}\right \rceil, 0,2^{b-1}\right)-z^{\prime}\right]
\label{equ3}
\end{equation}
\begin{equation}
s_{pow-2}=2^{\lfloor\log _{2} s\rceil}
\label{equ4}
\end{equation}

\begin{equation}
z^{\prime}=-\frac{{x}_{min}}{s_{pow-2}}
\label{equ5}
\end{equation}
where ${{x}_{min}}$ is the mapped smallest floating-point number in the vector ${\boldsymbol x}$. Equation (\ref{equ4}) is the Power-of-Two scale replaced by the naive method, and Equation (\ref{equ5}) is the zero-point updated after scale change.

The discussion on Power-of-Two scale can be divided into the following two cases:

\begin{itemize}
\item When ${s_{pow-2} < s}$, ${\lfloor\log _{2} s\rceil}$ rounded down to ${\lfloor\log _{2} s\rfloor}$, the grid step length decreases, resulting in more clipping error, i.e., there are more data clipped at the maximum number of fixed points, and the values of outliers all become the same value after the dequantization.
\item When ${s_{pow-2} > s}$, ${\lfloor\log _{2} s\rceil}$ rounded up to ${\lceil\log _{2} s\rceil}$, the grid step length increases, resulting in more rounding error, i.e., there are more data laying at every grid, and they all become a same number after dequantization.
\end{itemize}

To visualize the Power-of-Two scale factor caused clipping and rounding errors, we illustrate with specific quantization data distributions. We use 6-bit symmetric uniform affine quantization for the pre-training weights of the DarkNet53 \cite{redmon2018yolov3} on Hand dataset \cite{mittal2011hand}. Figure \ref{fig2} shows the histogram of the data distribution of the quantized weights at 73rd layer. The orange mask is the distribution of normal scale quantized data, and the blue histogram is the distribution of Power-of-Two scale quantized data. Figure \ref{fig2(a)} shows the data distribution using $2^{\lfloor\log _{2} s\rfloor}$ quantization, and Figure \ref{fig2(b)} shows the data distribution using $2^{\lceil\log _{2} s\rceil}$ quantization. The relationship between clipping error and rounding error is that they always trade with each other.

The selection of scale in Power-of-Two quantization is essentially a trade-off between rounding error and clipping error. The naive method simply finds a Power-of-Two scale with the smallest value difference from the original scale, which has no connection to model accuracy. Reducing the numerical difference between the Power-of-Two scale and the original scale in every layer does not necessarily decrease the task loss of the model. Besides, even using the greedy strategy to directly select the best Power-of-Two scale of a single layer often can only obtain the local optimal scale factor. 
So we think there should be a solider method, which adopts task loss as the criterion to choose Power-of-Two scale, theoretically trades off the clipping error and rounding error and eventually obtain the optimal solution of the whole network.

\subsection{Regression Loss Function for Reconstruction}
Recent excellent PTQ work AdaRound \cite{nagel2020up} performed a second-order Taylor expansion of difference between the task losses before and after quantization by two strong assumptions. It finally convert the difference to a L-2 loss of feature map before and after quantization between layers. \cite{li2020brecq} changed the assumptions of \cite{nagel2020up} and applied this theory to block reconstruction, eventually converting task loss to an L-2 loss minimization of feature map before and after quantization between blocks. Their crude assumptions bring crude conclusion, and the reconstructed regression loss functions all become L-2 loss functions. For L-2 loss minimization, it is easy to show that the regression value is actually the mean value of the array. The mean value is sensitive to outliers of the array, which means that using L-2 loss minimization will produce more rounding error than clipping error. If other reconstruction schemes are used instead of L-2 loss minimization, this comes back to the problem of trading off rounding and clipping errors mentioned in Sec 3.1.

\section{Method}
The two challenges mentioned in Sec 3.1 and Sec 3.2 have led to a collapse in model accuracy for Power-of-Two low-bit PTQ, a phenomenon that is more pronounced in light-weight networks like MobileNetV2. In this section, we propose two methods to rescue Power-of-Two PTQ from accuracy collapse. These two methods are theoretically well-founded and show significant performance improvement in practice.
\subsection{Power-of-Two Scale Group}
To address the first problem mentioned in Sec 3.1, we abandon the naive method of determining the Power-of-Two scale layer by layer and look for the Power-of-Two scale group of the entire network or block instead.

The goal of trading off rounding error and clipping error is to make the model more accurate, i.e., the model has a lower task loss. So we directly use the task loss as metric to evaluate performance of quantization.
\begin{equation}
\underset{\boldsymbol{\hat{w}}}{\arg\min}\mathbb{E}[\mathcal{L}(\boldsymbol{\hat{w}})] \qquad \boldsymbol{\hat{w}} \in \mathbb{D}_{Q}
\label{equ6}
\end{equation}
where $\mathcal{L}(\cdot)$ is the loss function of the model, ${\boldsymbol{\hat{w}}}$ represents the quantized model weights and $\mathbb{D}_{Q}$ is the discrete space of fixed-point numbers.
In QAT, this process is easier to optimize by stochastic gradient descent for updating quantization parameters and weights. But in case of PTQ, we can only calibrate the model with a small portion of dataset. To solve this problem, \cite{nagel2020up} degenerated the loss difference using a second-order Taylor expansion into the Equation (\ref{equ7}).
\begin{equation}
\begin{aligned}
\underset{\Delta \boldsymbol{w}}{\arg\min} \mathbb{E}[\mathcal{L}(\boldsymbol{x}, \boldsymbol{tgt}, \boldsymbol{w}+\Delta \boldsymbol{w})-\mathcal{L}(\boldsymbol{x}, \boldsymbol{tgt}, \boldsymbol{w})]\\
\approx \mathbb{E}\left[\frac{1}{2} \Delta \boldsymbol{w}_{(k)}^{\mathrm{T}} \boldsymbol{H}(\boldsymbol{w}_{(k)}) \Delta\boldsymbol{w}_{(k)}\right]
\end{aligned}
\label{equ7}
\end{equation}
where $\boldsymbol{x}$ is the input to the model, $\boldsymbol{tgt}$ is the ground truth, $\boldsymbol{w}$ is the original weight of the model. ${\Delta \boldsymbol{w}}$ is the perturbation brought by the model quantization to the model weights.

However, the calculation of the Hessian matrix is too complex and solving this problem is not allowed by the computing resources of PTQ. Thus, they assumed that layers are mutual-independent and the second-order derivatives of pre-activation are constant diagonal matrix. In this way, they transformed the problem into an L-2 loss minimization with layer-by-layer feature map reconstruction. Solving this problem only requires focusing on the current layer and solving each subproblem as shown in Equation (\ref{equ8}).
\begin{equation}
\underset{\Delta W_{(k) i,:}}{\arg\min} \mathbb{E}\left[\left(\Delta \boldsymbol{w}_{(k) i,:} x_{(k)}\right)^{2}\right]
\label{equ8}
\end{equation}
\cite{li2020brecq} generalized this work. They ignored inter-block dependencies, and used the diagonal Fisher information matrix (FIM) instead of the pre-activation Hessian matrix ~\cite{lecun2012efficient}. Our optimization objective can be converted into a block-by-block feature map reconstruction problem, as shown in Equation (\ref{equ9}).

\begin{equation}
\underset{\boldsymbol{\hat{w}}}{\arg \min } \mathbb{E}\left[\Delta \boldsymbol{y}_{(k)}^{\mathrm{T}} \text{diag}\left( \left(\frac{\partial L}{\partial \boldsymbol{y}_{(k)i}}\right)^{2} \right) \Delta \boldsymbol{y}_{(k)}\right]
\label{equ9}
\end{equation}
Where $ \Delta \boldsymbol{y}_{(k)}$ is the change of output by quantization. The middle term is the diagonal Fisher information matrix.

\begin{figure}[t]
\centering
\includegraphics[width=1\columnwidth]{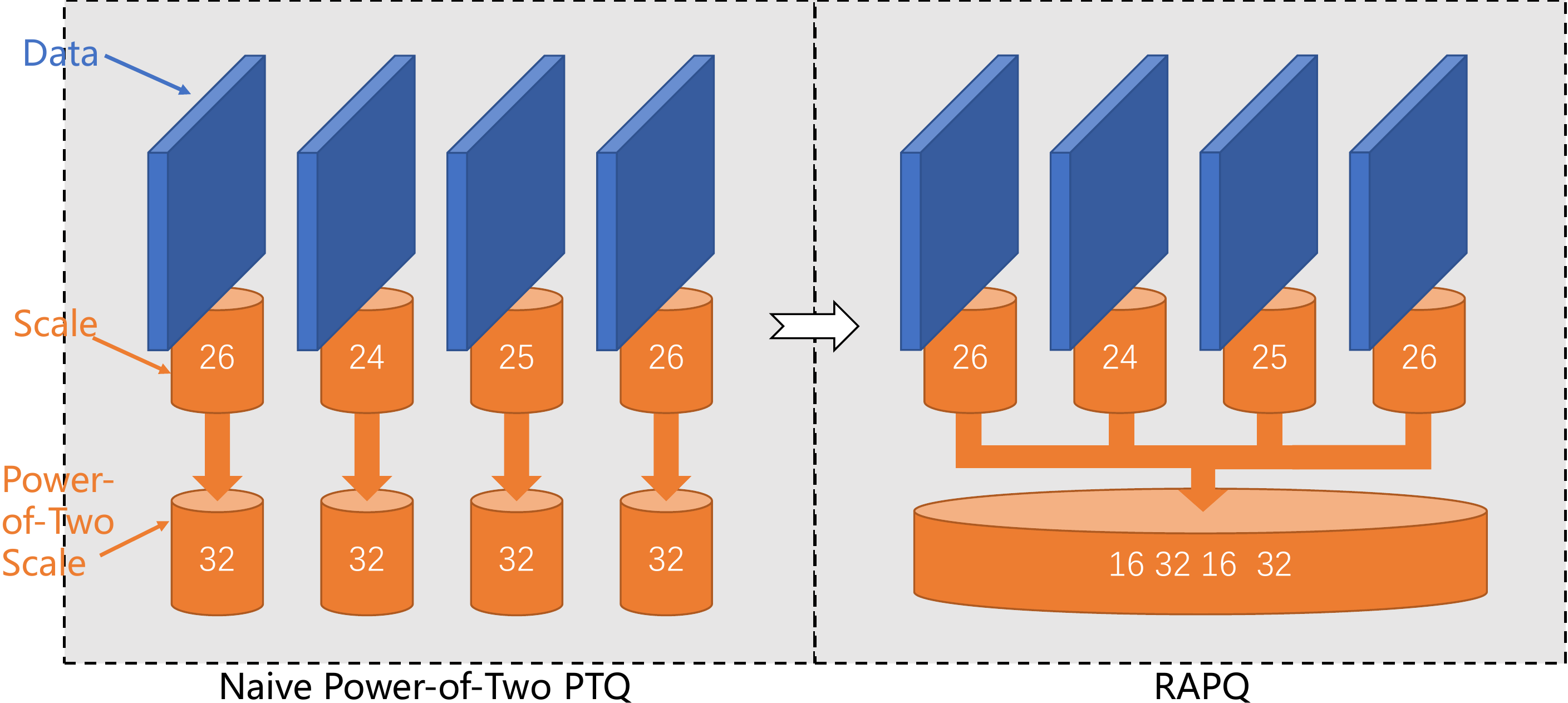} % Reduce the figure size so that it is slightly narrower than the column. Don't use precise values for figure width.This setup will avoid overfull boxes.
%0502\vspace{-3mm}
\caption{Method 4.1 Power-of-Two Scale Group}
\label{fig3}
\end{figure}
\setlength{\belowcaptionskip}{-5pt}

According to Equation (\ref{equ9}), if we want to calculate the optimal Power-of-Two scale and weights for the whole block, we no longer need to calculate the extremely complicated Hessian matrix, but still need to solve the NP-hard discrete optimization problem. Because the calculation of $ \Delta \boldsymbol{y}_{(k)}$ requires $\boldsymbol{\hat{y}_{(k)}}$ from the discrete quantization space. Therefore, we degenerate Equation (\ref{equ9}) to a continuous optimization problem Equation (\ref{equ10}) which is based on soft quantization variables, in order to solve it using the back propagation algorithm.
\begin{equation}
\begin{aligned}
\underset{\boldsymbol{U,V}}{\arg\min} \| \Delta \boldsymbol{y}_{(k)} \|_{F}^{2}+\lambda f_{reg}(\boldsymbol{U})+\mu f_{reg}(\boldsymbol{V})\\
=\left\|\boldsymbol{\widetilde{y}}_{(k)}-\boldsymbol{y}_{(k)}\right\|_{F}^{2}+\lambda f_{reg}(\boldsymbol{U})+\mu f_{reg}(\boldsymbol{V})
\end{aligned}
\label{equ10}
\end{equation}
where $\|\cdot\|_{F}^{2}$ denotes the L-2 loss and $\boldsymbol{\widetilde{y}_{(k)}}$ is the result of  $\boldsymbol{\widetilde{W}}$ and input calculation. $\boldsymbol{\widetilde{W}}$ is weights of the soft quantization. The so-called soft quantization is to first replace the discrete quantized variables with continuous float variables in order to back propagate and help the model converge. With the help of differentiable regularizer $\lambda f_{reg}(\boldsymbol{U})$ and $\mu f_{reg}(\boldsymbol{V})$, the data will eventually converge or clip to a truly quantized fixed-point discrete space. Expanding $\boldsymbol{\widetilde{W}}$, we have Equation (\ref{equ11}).
\begin{equation}
\begin{gathered}
\begin{footnotesize}
\boldsymbol{\widetilde{\boldsymbol{W}}}=\widetilde{s}_{pow-2}\cdot\left[ \operatorname{clip}\left(\left\lfloor \frac{\boldsymbol{W}}{\widetilde{s}_{pow-2}}+z^{\prime}\right\rfloor+h_{1}(\boldsymbol{U}), 0,2^{b-1}\right)-z^{\prime}\right]
\end{footnotesize}\\
\widetilde{s}_{pow-2}=2^{\left\lfloor\log_{2}s\right\rfloor+h_{2}(\boldsymbol{V})}\\
 z^{\prime}=-\frac{\boldsymbol{x}_{(\min) }}{\widetilde{s}_{pow-2}}=\frac{s\cdot z}{\widetilde{s}_{pow-2}}
\end{gathered}
\label{equ11}
\end{equation}
where $\left\lfloor\cdot\right\rfloor$is the downward rounding operation, $\boldsymbol{\widetilde{W}}$ is the weight of soft quantization, and ${\widetilde{s}_{pow-2}}$ is the soft Power-of-Two scale. $h_{1}(\cdot)$ and $h_{2}(\cdot)$ are differentiable functions that take values between $[0,1]$ and serve to process the trainable tensor $\boldsymbol{U}$,$\boldsymbol{V}$ in soft quantization and map them to 0 or 1 eventually.

However, we cannot guarantee that the trainable variables $\boldsymbol{U}$ of scale and $\boldsymbol{V}$ of weight in Equation (\ref{equ10}) converge at the same time. If $\boldsymbol{V}$ converges before $\boldsymbol{U}$ it will lead to the problem that the converged Power-of-Two scale does not match the converged weight. Therefore, we convert the nonlinear programming problem into two binary constrained optimization problems to be solved in two steps.

\begin{figure}[t]
\centering
\includegraphics[width=0.853\columnwidth]{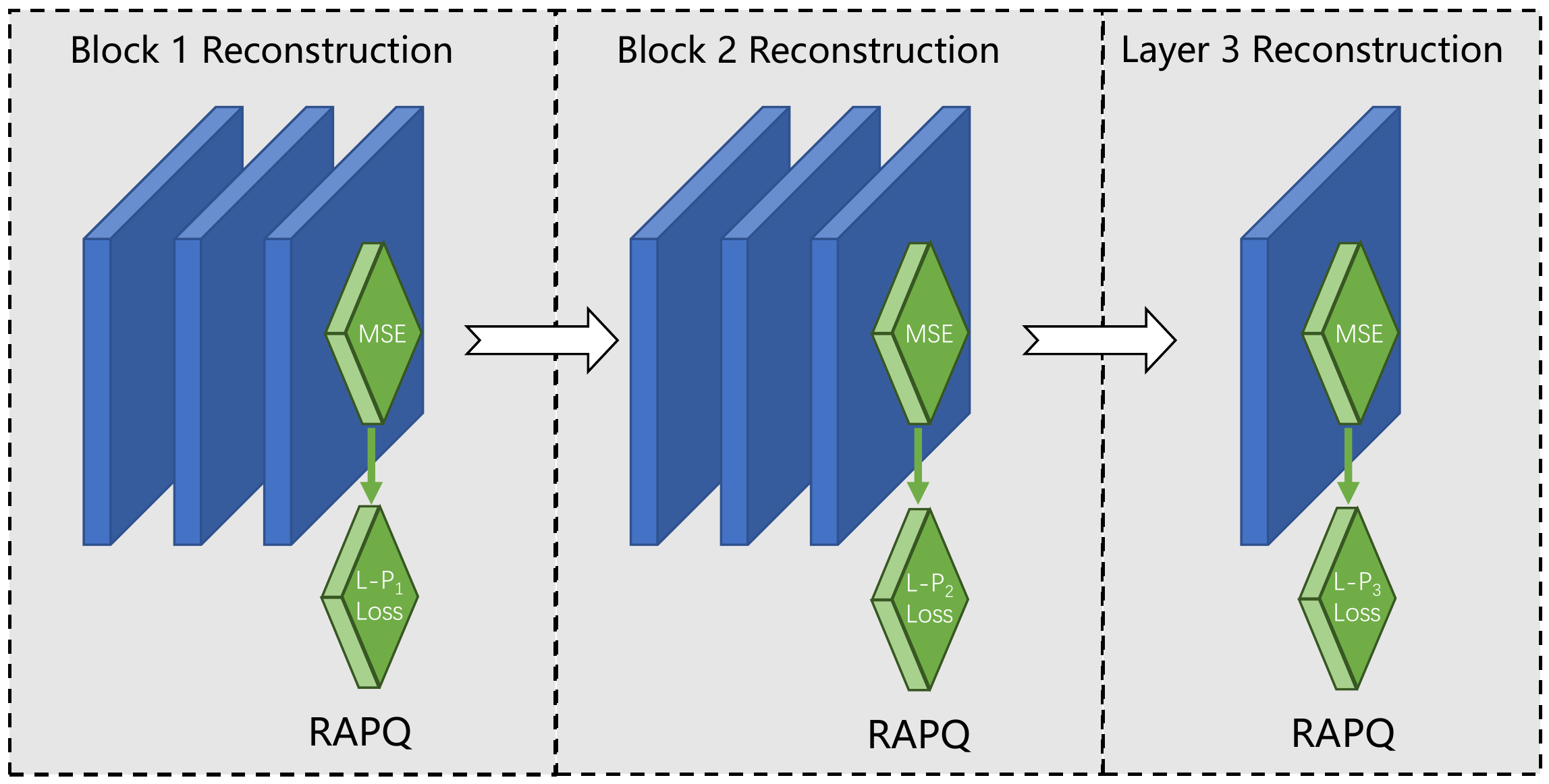} % Reduce the figure size so that it is slightly narrower than the column. Don't use precise values for figure width.This setup will avoid overfull boxes.
%0502\vspace{-3mm}
\caption{Method 4.2 BN based L-P Loss}
\label{fig4}
\end{figure}
\setlength{\belowcaptionskip}{-5pt}

\begin{enumerate}
    \item Look for the optimal solution for the Power-of-Two scale group by Equation(\ref{equ12}).
\begin{equation}
\underset{\boldsymbol{V}}{\arg\min}\left\|\boldsymbol{\widetilde{y}}_{(k)}-\boldsymbol{y}_{(k)}\right\|_{F}^{2}+\mu f_{reg}(\boldsymbol{V})
\label{equ12}
\end{equation}
Freeze the Power-of-Two scale after the variable $\boldsymbol{V}$ converges.
    \item Look for the optimal solution to the quantized weight $\boldsymbol{\widetilde{W}}$ by Equation (\ref{equ13}).
\begin{equation}
\underset{\boldsymbol{U}}{\arg\min}\left\|\boldsymbol{\widetilde{y}}_{(k)}-\boldsymbol{y}_{(k)}\right\|_{F}^{2}+\lambda f_{reg}(\boldsymbol{U})
\label{equ13}
\end{equation}
After convergence of the variable $\boldsymbol{U}$, the quantized  $\boldsymbol{\hat{W}}$ are stored.
\end{enumerate}
Thus problem (11) is transformed into two binary constrained optimization problems. Both problems are large scale combinatorial problems, and referring to the work of \cite{nagel2020up} we solve these two problems using an efficient approximation algorithm Hopfield methods \cite{hopfield1985neural}.

For the training functions $h_{1}(\cdot)$, $h_{2}(\cdot)$ we adopt the rectified sigmoid function, as shown in Equation (\ref{equ14}), which is proposed in \cite{louizos2018learning}.
\begin{equation}
h\left(\boldsymbol{x}_{i j}\right)=\operatorname{clip}\left(\operatorname{sigmoid}\left(\boldsymbol{x}_{i j}\right)(\xi-\gamma)+\gamma, 0,1\right)
\label{equ14}
\end{equation}
where $\xi$ and $\gamma$ are the stretching parameters, fixed at 1.1 and -0.1, respectively. They help the rectified sigmoid function to converge more easily to the extremities 0 and 1, and are not as prone to gradient vanishing as sigmoid function. For regularizer we choose:
\begin{equation}
f_{reg}(\boldsymbol{x})=\sum_{i, j} 1-\left|2 h\left(\boldsymbol{x}_{i, j}\right)-1\right|^{ \beta}
\label{equ15}
\end{equation}
Because in this regularizer, we can achieve annealing by controlling the ${ \beta}$. $h(x)$ can easily converge to 0 or 1 when the value of ${ \beta}$ drops to a low value.

The activations cannot be quantized using adaptive rounding because they vary with different input. Thus, we can only adjust its zero-point and Pow-of-Two scale. Referring to back propagation of TQT~\cite{jain2019trained}, when calculating the gradient, we approximate by taking $s \approx 2^{\left\lfloor\log _{2} s \right\rceil} $and $\left\lfloor\frac{\boldsymbol{x}}{s}+z\right\rceil \approx \frac{\boldsymbol{x}}{s}+z$. Then Power-of-Two scale $s_{pow-2}$ can be back-propagated by Equation (\ref{equ16}).

\begin{equation}
\begin{split}
&\nabla_{{{log}_{\boldsymbol2}}{S_{pow-2} }} \boldsymbol{\hat{x}}=\\
&s\cdot\ln 2
\left\{\begin{array}{c}
\begin{aligned}
&\lfloor\tfrac{\boldsymbol{x}}{s}+z\rceil-(\tfrac{\boldsymbol{x}}{s}+z)
&& 0\le \lfloor \tfrac{\boldsymbol{x}}{s}+z \rceil \le 2^b-1 \\ 
&0 
&&\lfloor \tfrac{\boldsymbol{x}}{s}+z \rceil < 0 \\ 
&2^{b-1} 
&& \lfloor \tfrac{\boldsymbol{x}}{s}+z \rceil > 2^b-1
\end{aligned}
\end{array}
\right.
\end{split}
\label{equ16}
\end{equation}
\subsection{BN-based L-P Loss}
To address the second problem mentioned in Sec 3.2, we introduce the minimum L-P loss problem, as shown in Equation (\ref{equ17}), to measure the difference between the feature maps before and after quantization.
\begin{equation}
\underset{f}{\arg\min} \mathcal{L}(f)=\underset{f}{\arg\min} \sum_{i=1}^{n}\left|\boldsymbol{x}_{i}-f(a, b, c, \cdots)\right|^{P}
\label{equ17}
\end{equation}
where $P\in [1,+\infty)$.There is the following conclusions:
\begin{itemize}
\item L-1 loss is ${Median}$ regression. It is not sensitive to outliers since ${Median}$ simply has an equal number of positive and negative deviations.
\item L-2 loss is ${Mean}$ regression, which is more sensitive to outliers since it has a positive and negative deviation of the sum of 0.
\item L-$+\infty$ loss is ${Midrange}$ ${Number}$ regression, highly sensitive to outliers, with a maximum positive deviation and a minimum negative deviation summed to 0.
\end{itemize}
From these three special cases it is easy to see a proven theory that the sensitivity of the regression value of L-P loss to outliers increases as the P-value increases.

It is not reasonable to use one L-P loss for all of layers because the data distribution of each layer activation is very different. For example, when quantizing the last layer, we should keep his outlier characteristics because these outliers are about to become the most important scores to predict the classification; when quantizing the middle layer, if the regression is insensitive to outliers, we can let the data distribution before and after quantization more similarly.

As shown in Equation (\ref{equ18}), the batch nomlization (BN) \cite{ioffe2015batch} is widely available in today's neural networks. 
\begin{equation}
\widetilde{y}_{(k)}=\gamma \cdot \frac{y_{(k)}-\mu_{B}}{\sqrt{\sigma_{B}^{2}+\varepsilon}}+\beta 
\label{equ18}
\end{equation}
where ${y}_{(k)}$ is the original activation of the layer, and $\widetilde{y}_{(k)}$ is the activation after BN. $\mu_{B}$,$\sigma_{B}^{2}$ are the Mean and Variance of the mini-batch while training. 
When the model stops training, $\mu_{B}$ and $\sigma_{B}^{2}$ are replaced by $\mu_{running}$ and $\sigma_{running}^{2}$,which are the running Mean and running Variance of the statistics in training.
 $\gamma$ and $\sigma_{running}^{2}$ reflect, to some extent, the statistical Variance of the data when the model was previously trained. The Variance is used to measure the degree of deviation between a random variable and its mathematical expectation and can reflect the degree of deviation of the data.

$\gamma$ reflects the Variance information of the data after BN, while $\sigma_{running}^{2}$ reflects the Variance information of the data before BN. In practical hardware inference, the BN-layer is fused in the convolutional layer, so we need the Variance information after BN.
Taking block as the unit, the layers that do not belong to block take each layer as a single unit. We take the parameter $\gamma$ of BN information of the last layer of each unit, and find the average $\gamma$ of all channels as the deviation degree flag of that layer. We define the formula for calculating the P-value of ${k}$-th layer BN-based L-P loss:
\begin{equation}
P_{(k)}=1+\alpha \cdot \operatorname{sigmoid}\left(\frac{1}{N_{k}} \sum_{i=1}^{N_{k}} {\gamma}_{(k)j}-\beta\right)
\label{equ19}
\end{equation}
where $\alpha$, $\beta$ are two adjustable parameters,$\alpha \in (0,1]$, $\beta \in \mathbb{R}$. In order to satisfy the L-P norm definition, $P>1$ is necessary, but the model is difficult to converge and prone to large clipping errors when $P>2$. Therefore, we introduce such a differentiable function with a value domain of $[1,2]$. The larger P-value is, the harder model converges. But when the number of iterations is sufficient, we can increase the disparity of P-values by turning up the $\alpha$. When the model has extreme $\gamma$ values, $\beta$ is needed to help reduce the problem of gradient vanishing of the sigmoid function. For ease of calculation, we usually keep the P-value to 1~2 decimal places.

\begin{algorithm}[tb]
\caption{: RAPQ Power-of-Two Quantization}
\label{alg:algorithm}
\textbf{Input}: Calibration dataset, Pretrained FP model\\
\textbf{Parameter}: ${I}_{s}$, ${I}_{w}$, ${I}_{a}$ iterations\\
\textbf{Output}:Quantized Model
\begin{algorithmic}[1] %[1] enables line numbers
\STATE Init s and z by MSE, $[P_i]$ by Equation(\ref{equ19})
\WHILE{ $i = 1,2,\cdots, $N-th}
\IF {weight optimization in warm-up}
\STATE ${I}_{s}$ iterations to find ${s}_{pow-2}$ group of weight
\ELSE
\STATE Freeze ${s}_{pow-2}$. ${I}_{w}$ iterations to optimize $\hat{W}$
\ENDIF
\ENDWHILE\\
\WHILE{$i = 1,2,\cdots, $N-th}
\STATE ${I}_{a}$ iterations to find ${s}_{pow-2}$ group of activation
\ENDWHILE
\STATE \textbf{return} Power-of-Two Quantized Model
\end{algorithmic}
\end{algorithm}

\section{Experiments}
It is widely recognized by peers that for CNN-CV tasks, excellent results of quantization method on classification task models are the basis for awesome results of other kind of task models. So we have conducted extensive experiments based on ImageNet \cite{russakovsky2015imagenet} dataset to demonstrate the superiority of our method.
We randomly pick a total of 1024 images for PTQ calibration in each experiment. To be fair, we optimized each model with 80,000 weight iterations and 5000 activation iterations in order to fully converge. At this point, the parameter 
$\alpha$ corresponding to Equation (\ref{equ19}) is set to 0.9 and $\beta$ is set to 1. Although it is not listed in the table, we have confirmed that we can achieve better results than the data in the table if we fine-tune the $\alpha$,$\beta$ in Equation (\ref{equ19}) for different models. The optimization of the Power-of-Two scale group will be done during the weight warm-up process.
This section is divided into four parts. The first part is an ablation study to demonstrate the effectiveness of our two methods. The second part is a comparison with Power-of-Two quantization work, which demonstrates that our work achieves SOTA. The third part is a comparison with SOTA PTQ work, which demonstrates that we can still achieve a performance close to that of other unconstrained quantization work while satisfying the constraint of hardware-friendly property.
The fourth part is set for the time-limited scenario. In order to obtain better quantization results quickly, then the L-P loss and number of iterations setting in the fourth part can be used.

\begin{table*}[!t]
\scalebox{0.7}
% \footnotesize
%     \caption{Expriment 5.2 Comparison with other Power-of-Two}\label{tab_imagenet_weight}
    \centering
    \begin{adjustbox}{max width=\textwidth}
    \begin{tabular}{lrrrrrrr}
    \toprule
    \textbf{Methods} & \textbf{Bits (W/A)} & \textbf{ResNet-18} & \textbf{ResNet-50} & \textbf{MobileNetV2} & \textbf{RegNet-600MF} & \textbf{RegNet-3.2GF}\\
    \midrule
    Full Prec. & 32/32 & 71.08 & 77.00 & 72.49 & 73.71 & 78.36\\
    \midrule

    % DorefaNet*~\cite{zhou2016dorefa} & 4/4 & 69.2 & 75.1 & 64.33 & - & - \\
    TQT~\cite{jain2019trained} (QAT) & 4/8 & 67.90* & 74.40 & 47.16 & - & - \\
    \midrule
    APoT~\cite{li2019additive} (QAT) & 4/4 & \textbf{70.70} & \textbf{76.60} & - & - & - \\
    \textsc{RAPQ}~(Ours) & 4/4 & {69.28} & {74.64} & \textbf{64.48} & \textbf{69.59} & \textbf{74.25}\\
    \midrule
        \textsc{RAPQ}~(Ours) & 2/4 & 65.32 & 69.71 & \textbf{48.12} & \textbf{61.48} & \textbf{69.49}\\
    \bottomrule
    \end{tabular}
    \end{adjustbox}
    
%0502\vspace{-3mm}

\caption{Comparison of RAPQ and SOTA Power-of-Two quantization work on ImageNet}
\label{Expriment 5.2}
\end{table*}
% \vspace{-3mm}

\begin{table*}[!t]
\scalebox{0.7}
% \footnotesize
%     \caption{Expriment 5.3 Comparison with other sota PTQ}\label{tab_imagenet_weight}
    \centering
    \begin{adjustbox}{max width=\textwidth}
    \begin{tabular}{lrrrrrrr}
    \toprule
    \textbf{Methods} & \textbf{Bits (W/A)} & \textbf{ResNet-18} & \textbf{ResNet-50} & \textbf{MobileNetV2} & \textbf{RegNet-600MF} & \textbf{RegNet-3.2GF}\\
    \midrule
    Full Prec. & 32/32 & 71.08 & 77.00 & 72.49 & 73.71 & 78.36\\
    \midrule
    ACIQ~\cite{banner2019aciq} & 4/4 & 67.00 & 73.80 & - & - & -\\
    ZeroQ~\cite{cai2020zeroq}* & 4/4 & 20.80 & 4.27 & 25.24 & 27.95 & 12.38 \\
    AdaRound~\cite{nagel2020up}* & 4/4 & 68.45 & 74.51 & 63.94 & - & - \\
    AdaQuant~\cite{hubara2020adaquant} & 4/4 & \textbf{69.60} & \textbf{75.90} & 44.52* & - & - \\
    Bit-Split~\cite{wang2020bitsplit} & 4/4 & 67.56 & 73.71 & - & - & - \\
    {Brecq}~\cite{li2020brecq} & 4/4 & \textbf{69.60} & 75.05 & \textbf{66.57} & 68.33 & 74.21 \\
    \textsc{RAPQ}~(Ours) & 4/4 & 69.28 & \ 74.64 & 64.48 & \textbf{69.59} & \textbf{74.25}\\
    \midrule
    ZeroQ~\cite{cai2020zeroq}* & 2/4 & 0.10 & 0.11 & 0.13 & 0.07 & 0.06 \\
    AdaRound~\cite{nagel2020up}* & 2/4 & 64.14 & 68.40 & 41.52 & 59.27 & 65.33 \\
    AdaQuant~\cite{hubara2020adaquant}* & 2/4 & 0.16 & 0.19 & 0.08 & 0.10 & 0.11 \\
    {Brecq}~\cite{li2020brecq}& 2/4 & 64.80 & \textbf{70.29} & \textbf{53.34} & 59.31 & 67.15\\
        \textsc{RAPQ}~(Ours) & 2/4 & \textbf{65.32} & 69.71 & 48.12 & \textbf{61.48} & \textbf{69.40}\\
    \bottomrule
    \end{tabular}
    \end{adjustbox}
%0502\vspace{-3mm}
\caption{Comparison of RAPQ and SOTA PTQ work on ImageNet}
\label{Expriment 5.3}
\end{table*}

\subsection{Ablation Study}
It is generally accepted that MobileNetV2 \cite{sandler2018mobilenetv2} is one of the most difficult lightweight networks to quantize because its weights are extremely susceptible to perturbations.
To show the superiority of our method, we conduct ImageNet experiments using MobileNetV2.
As shown in Table \ref{Expriment 5.1}, we perform ablation study limited with weight IN2 activation INT4( W2/A4 ).
It is easy to see that Power-of-Two scale group(Po2 SG) method rescues the accuracy of MobileNetV2. BN-based L-P loss can further improve their accuracy.
\begin{table}[h]
%     \caption{Ablation study}\label{tab_imagenet_weight}
\centering
\begin{tabular}{ccc}
\toprule
Model  & MobileNetV2\\
\midrule
Naive Power-of-Two       & 2.64 \\
Po2 SG    & 46.48       \\
Po2 SG + BN-based L-P loss   & 48.12      \\
\bottomrule
\end{tabular}
%0502\vspace{-3mm}
\caption{Ablation study}
\label{Expriment 5.1}
\end{table}

\begin{table}
% \footnotesize
%     \caption{RAPQ Quick Mode}\label{tab_imagenet_weight}
\centering
\begin{tabular}{lrr}
\toprule
Model  & W2/A4 Acc & W4/A4 Acc\\
\midrule
MobileNetV2       & 46.68  & 62.55\\
ResNet-18    & 64.76   & 69.26    \\
ResNet-50   & 69.20   & 74.53  \\
RegNet-600MF   & 60.86 & 69.51      \\
RegNet-3.2GF   & 68.47 &  74.39   \\
\bottomrule
\end{tabular}
%0502\vspace{-3mm}
\caption{The experiment results of RAPQ Quick Mode on ImageNet}
\label{Expriment 5.4}
\end{table}

\subsection{Comparison with SOTA Power-of-Two}
As shown in Table \ref{Expriment 5.2}, we compared our experiment results with TQT and APoT. TQT uses a uniform affine QAT scheme and APoT uses a non-uniform affine QAT scheme while we use a uniform affine PTQ scheme. Both TQT and APoT need the whole dataset while we need only 1024 of it. They spend more than 10 times as long as we do to quantize the model. APoT has achieved better results with W4/A4 than we have on ResNet \cite{he2016deep}, but it introduces weight normalization to smooth the learning process of clipping range in weight. It’s impossible to incorporate this technique with BN folding so that it can only reproduce in academic setting\cite{li2021mqbench}. We are the first in the Power-of-Two quantization work to achieve quantization for MobileNetV2 with W2/A4.

\subsection{Comparison with SOTA PTQ}
As Table \ref{Expriment 5.3} shows, we compare our hardware-constrained PTQ work with the PTQ work without constraint. Our experiment results on RegNet \cite{radosavovic2020regnet} are better than those of SOTA PTQ without hardware constraint, thanks mainly to Method 4.2, because the performance of Power-of-Two scale is worse than that of ordinary scale.
    
\subsection{Quick Mode}
Many application scenarios are not extreme in terms of accuracy requirement, and they value shorter quantization time. For such scenarios, we introduce a Quick Mode with only 20,000 weight iterations and 1,000 activation iterations. Correspondingly, the parameter $\alpha$ in Equation (\ref{equ19}) is set to 0.1 and $\beta$ is set to 1. This scheme takes only 10 minutes to quantize Resnet18 with Intel i9-10980XE + Nvidia RTX3090. As shown in Table \ref{Expriment 5.4}, although it takes very short time, it also has a notable accuracy performance.

\section{Conclusion}
In this paper, we propose RAPQ, a Power-of-Two low-bit post-training quantization framework. At first, we analyze the reasons for the accuracy collapse of Power-of-Two PTQ, i.e., the failure to theoretically trade off rounding error and clipping error and the rough setting of the regression loss while reconstructing.
For the first reason we propose a method for finding Power-of-Two scale group of CNN model. For the second reason we propose a method to formulate the regression loss based on the BN information of each unit. The experiments show that our work not only reaches SOTA in the field of Power-of-Two quantization, but also does not fall short of other unconstrained quantization methods. When quantizing MobileNetV2 with W2/A4, our work can achieve an accuracy of 48\%, which was not achieved by all previous work on Power-of-Two quantization (including QAT).

\section*{Acknowledgments}

This work was supported by the National Key Research and Development Program of China (Grant No. 2018YFE0203801).

\bibliographystyle{named}
\bibliography{ijcai22}

\end{document}